\documentclass[10pt,twocolumn,letterpaper]{article}

\usepackage[lined,boxed,commentsnumbered,ruled]{algorithm2e}
\usepackage{iccv}
\usepackage{times}
\usepackage{epsfig}
\usepackage{graphicx}
\usepackage{amsmath}
\usepackage{amssymb}
\usepackage{bm}
\usepackage{tabu}
\usepackage{float}
\usepackage{subfigure}
\usepackage{amsmath}
\usepackage{tablefootnote}

\usepackage[breaklinks=true,bookmarks=false]{hyperref}

\iccvfinalcopy 

\ificcvfinal\fi

\begin{document}

\title{Taking A Closer Look at Visual Relation : \\ Unbiased Video Scene Graph Generation with Decoupled Label Learning}

\author{Wenqing Wang\textsuperscript{1} \quad Yawei Luo\textsuperscript{1}\thanks{Corresponding author}{} \quad Zhiqing Chen\textsuperscript{1} \quad Tao Jiang\textsuperscript{1} \quad Lei Chen\textsuperscript{2} \quad Yi Yang\textsuperscript{1} \quad Jun Xiao\textsuperscript{1}\\
\textsuperscript{1}Zhejiang University \quad
\textsuperscript{2}FinVolution Group\\
}

\maketitle
\ificcvfinal\thispagestyle{empty}\fi

\begin{abstract}
Current video-based scene graph generation (VidSGG) methods have been found to perform poorly on predicting predicates that are less represented due to the inherent biased distribution in the training data. In this paper, we take a closer look at the predicates and identify that most visual relations (\eg \texttt{sit\_above}) involve both actional pattern (\texttt{sit}) and spatial pattern (\texttt{above}), while the distribution bias is much less severe at the pattern level. Based on this insight, we propose a decoupled label learning (DLL) paradigm to address the intractable visual relation prediction from the pattern-level perspective. Specifically, DLL decouples the predicate labels and adopts separate classifiers to learn actional and spatial patterns respectively. The patterns are then combined and mapped back to the predicate. Moreover, we propose a knowledge-level label decoupling method to transfer non-target knowledge from head predicates to tail predicates within the same pattern to calibrate the distribution of tail classes.  We validate the effectiveness of DLL on the commonly used VidSGG benchmark, \ie VidVRD. Extensive experiments demonstrate that the DLL offers a remarkably simple but highly effective solution to the long-tailed problem, achieving the state-of-the-art VidSGG performance. 
\end{abstract}

\section{Introduction}

Video-based scene graph generation (VidSGG) aims to represent video content as dynamic graphs constructed by \texttt{$\langle$subject, predicate, object$\rangle$} triplets. It offers high-level understanding and summarization of video knowledge, which can benefit downstream tasks such as visual question answering~\cite{antol2015vqa,tapaswi2016movieqa,xiao2021next}, video captioning~\cite{xu2015show}, and video retrieval~\cite{snoek2009concept,dong2021dual,wei2019neural}. Compared to its image-based counterpart ImgSGG~\cite{zellers2018neural,misra2016seeing}, VidSGG is considered a more challenging task, as the pairwise relations between visual entities are dynamic along the temporal dimension, making VidSGG a typical multi-label problem. Despite the vast body of literature on ImgSGG~\cite{zellers2018neural,misra2016seeing}, these characteristics prevent ImgSGG methods from being trivially applied to VidSGG. VidSGG remains a relatively under-explored problem with several unsolved issues at present.

\begin{figure}
\begin{center}
\includegraphics[width=\linewidth]{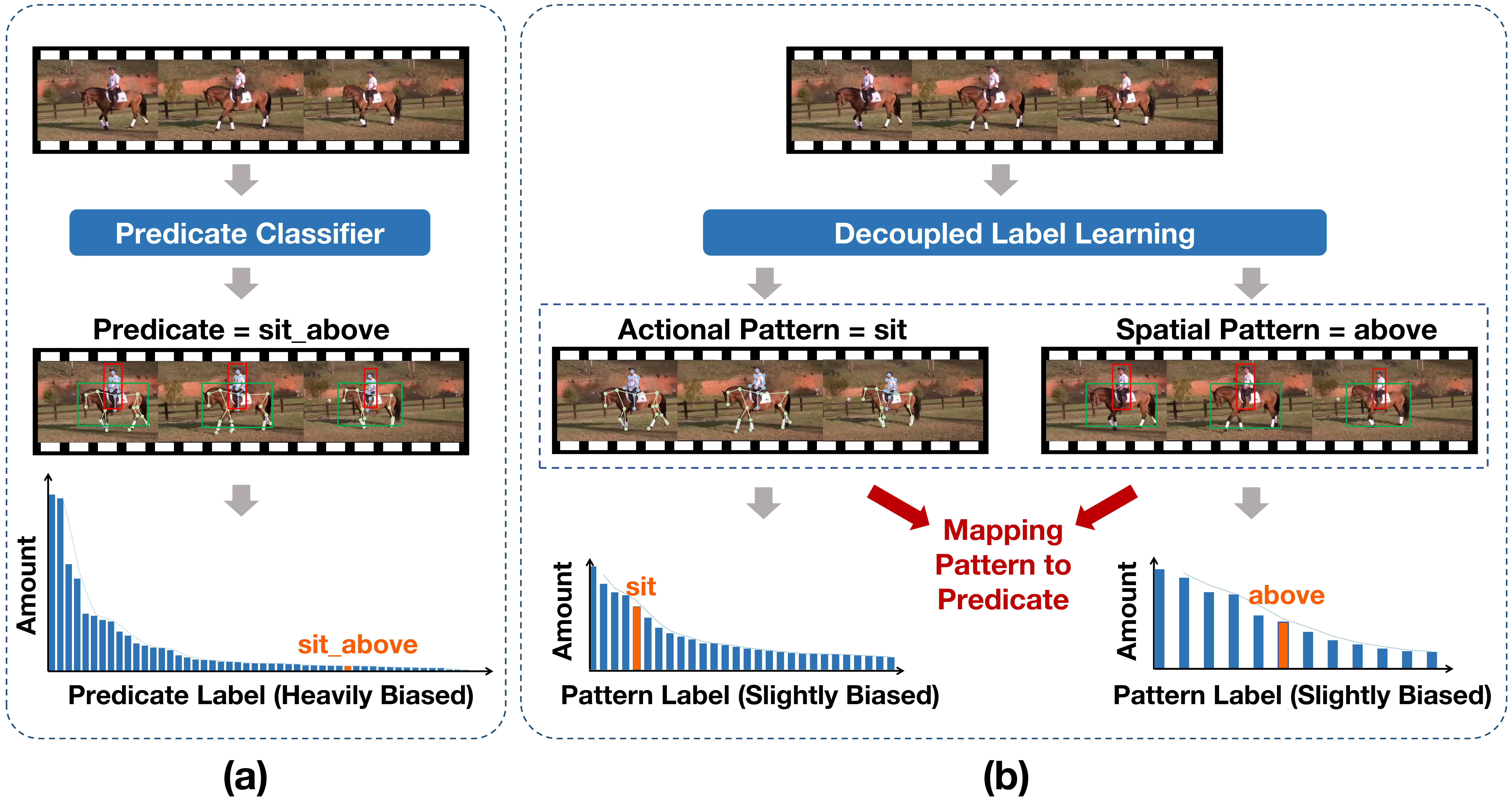} 
\end{center}
    \caption{(a) Previous canonical visual relation prediction pipeline. (b) A new DLL paradigm, which transforms the vanilla relation prediction task into a pattern-level problem.}
    \label{fig:introduction}
\vspace{-0.2cm}
 \end{figure}

Several recent attempts have been made to solve VidSGG by exploiting the spatio-temporal information of the video~\cite{qian2019video,liu2020beyond,teng2021target,cong2021spatial}. While these attempts have achieved some progress in pursuing overall performance by extracting short- and long-term information, they have ignored the inherent long-tailed nature of the data, leading to severely biased predicate predictions in the final results. For example, taking the commonly used VidVRD~\cite{shang2017video} dataset as an illustration, the head predicate categories nearly dominate the ground-truth annotations. Furthermore, even among all correctly detected predicates, a small number of head categories account for most of the overall performance. Additionally, the missing labels of VidSGG are inevitable during data annotation due to the fleeting temporal interaction or inconspicuous spatial relation of the objects. Compared to the head predicates, tail samples are more likely to be ignored by the annotators, further deteriorating the predicate bias in a video.

More recently, a few methods have noticed this phenomenon and endeavored to debias the visual relation. Li~\emph{et al.}\cite{li2021interventional} proposed a causality-inspired interaction to weaken the false correlation between input data and predicate labels. Xu \emph{et al.}~\cite{xu2022meta} considered temporal, spatial, and object biases in a meta-learning paradigm. These implicit approaches mitigate the long-tail problem to some extent, but the performance of tail classes is still unsatisfactory.

In this paper, we propose a more explicit method to tackle the biased VidSGG problem from the decoupled learning perspective. By taking a closer look at the visual relations in the video, we identify that most visual relations (\eg \texttt{sit\_above}) involve both actional pattern (\texttt{sit}) and spatial pattern (\texttt{above}). Surprisingly, compared to the original predicate-level label, the distribution bias is much less severe at the pattern level. Based on this insight, we propose a decoupled label learning (DLL) paradigm to transform the intractable visual relation detection into a pattern-wise prediction problem. Specifically, DLL employs an adversarial learning strategy to decouple the original video features into actional and spatial ones and then forwards them to the actional and spatial classifiers separately, in order to learn the decoupled patterns. The output patterns are then aggregated and mapped back to predicates.

To further boost the performance on tail visual relations, we propose a knowledge-level label decoupling method to calibrate the distribution of the preliminary predicates from pattern aggregation. We assume that head predicates are well represented with sufficient training data and the predicates within the same pattern ought to contain similar non-target knowledge  (\ie, correlation with other predicates). Motivated by this thought, we propose to decouple label knowledge into target and non-target knowledge and calibrate the non-target knowledge of tail classes with the help of head classes that are within the same pattern. To this end, we build a learnable predicate correlation matrix and align the tail prediction distribution to that of head classes. Our knowledge-level label decoupling method imposes tail classes to learn more from non-target knowledge of the head classes that are more abundant and calibrated.

In summary, this paper makes following contributions:
\begin{itemize}
    
    \item We propose the Decoupled Label Learning (DLL) method for VidSGG, which is the first attempt to address the intractable biased predicate prediction from a pattern-level perspective.

    \item We present a knowledge-level label decoupling approach to further boost the performance on tail visual relations, which calibrates the tail distribution using the non-target knowledge of head predicates.

    \item We conduct extensive experiments on a widely used VidSGG dataset: VidVRD~\cite{shang2017video}. The results demonstrate that DLL achieves state-of-the-art performance on various metrics across different scenarios, especially on those tail predicates. 
    
\end{itemize}

\section{Related Work}

\subsection{Image-based Scene Graph Generation}

ImgSGG has been an intensively studied task. Existing works can be roughly divided into two groups: 1) \textbf{Two-stage methods}~\cite{misra2016seeing,zellers2018neural,tang2019learning,chen2019knowledge}: conducting object detection and relationship prediction within two stages. 2) \textbf{One-stage methods}~\cite{liu2021fully,newell2017pixels,liao2020ppdm}: dealing with the object and relationship simultaneously in one go. Because of the long-tail problem, the performances of the traditional methods are far from satisfactory. Zeller~\emph{et al.}~\cite{zellers2018neural} first pointed out the imbalance of predicates in ImgSGG dataset. Chen~\emph{et al.}~\cite{chen2019knowledge} and Tang~\emph{et al.}~\cite{tang2019learning} also noticed this problem and proposed a new metric to measure the average performance~\emph{i.e.}, mean recall@K. Along this vein, various methods~\cite{han2022dbiased} have been proposed to solve biased relationship prediction, including depolarization strategies such as resampling~\cite{li2021bipartite}, reweighting~\cite{yan2020pcpl}, unbiased representation from bias~\cite{tang2020unbiased}, \emph{etc}. 

\subsection{Video-based Scene Graph Generation}

Generally, video-based and image-based SGG share a similar goal of detecting the visual objects and relationships in given data. Nevertheless, employing existing ImgSGG approaches straightly to parse a video is non-trivial due to the multi-label and dynamic nature of VidSGG. The existing VidSGG work can be roughly divided into two groups according to the format of dataset annotation:

\noindent{\textbf{Frame-based VidSGG:}} Ji~\emph{et al.}~\cite{ji2020action} proposed the first large-scale frame-level data set named \textbf{Action Genome (AG)}. Following this seminal work, Cong~\emph{et al.}~\cite{cong2021spatial} proposed a new network structure called ``space-time converter (STTran)'', which consists of a space encoder and a time decoder. Chen~\emph{et al.}~\cite{chen2023video} proposed a novel method for weakly-supervised task with only single-frame weak supervision and to generate pseudo labels for unannotated frames. Both of the methods are designed for better capturing the spatio-temporal context information for relationship recognition. 

\noindent{\textbf{Tracklet-based VidSGG:}} Shang~\emph{et al.} proposed a video-level dataset called \textbf{VidVRD}~\cite{shang2017video}, which aims to detect all visual relationship instances in the video in the form of relational triplets \texttt{$\langle$subject, predicate, object$\rangle$} and object tracks. In VidVRD~\cite{shang2017video}, the authors also proposed a widely used three-stage segment-based detection framework. However, this method is not ideal for long video detection. To remedy this, Feng~\emph{et al.}~\cite{feng2021exploiting} proposed a detection trajectory recognition paradigm by constructing consistent long-term object trajectories from videos, and then using transducers to capture the dynamic and visual relationships of objects. Besides, Liu~\emph{et al.}~\cite{liu2020beyond} also proposed a new sliding window scheme called ``VRD-STGC'' to simultaneously predict short- and long-term relations. Instead of preview segment-based and window-based methods, Woo~\emph{et al.}~\cite{woo2021and} presented a time span-suggested network ``TSPN'' to determine what and when to look in task. More recently, Shang~\emph{et al.}~\cite{shang2021video} proposed ``VidVRD-II'', which achieves iterative relational reasoning and joint relation classification. Gao~\emph{et al.}~\cite{gao2023compositional} first proposed a compositional and motion-based relation prompt learning framework (RePro) in open-vocabulary VidVRD setting. Albeit with these prior arts, only a few work has realized the long-tail predicate distribution as the bottleneck issue for VidSGG task~\cite{li2021interventional, xu2022meta}. 

Our work belongs to this tracklet-based vein. We adopt VidVRD-II~\cite{shang2021video} and VRD-STGC~\cite{liu2020beyond} as our baselines. Different from above methods, we target at long-tail predicate distribution problem and make pioneering attempt to address it from the perspective of decoupled label learning.

\subsection{Disentangled Representation Learning}
Disentangled Representation Learning (DRL) is an unsupervised learning technique aimed at identifying and separating the underlying factors of variation in observable data into representation form. The process of disentangling these factors into semantically meaningful variables aids in learning explainable data representations, imitating the meaningful reasoning process of humans. In recent years, numerous disentangling methods based on antagonism to decouple features have emerged in different fields, such as DADA,  proposed by Peng \emph{et al.}~\cite{2019Domain}, which simultaneously disentangles domain-invariant, domain-specific, and class-irrelevant features. The disentanglement process is carried out in an adversarial manner, with the disentangler generating features that deceive the class identifier trained on the labeled source domain. In the field of representation learning, disentangled representation learning has been heavily studied~\cite{desjardins2012disentangling,esmaeili2019structured,xiang2019disentangling}, which involves assigning different factors of variation to distinct dimensions of representation vectors. Jozsef Nemeth \emph{et al.}~\cite{2020Adversarial} also proposed an adversarial decoupling method based on group observation to separate content and style-related attributes. Yang \emph{et al.}~\cite{yang2022speech} used a gradient reversal layer (GRL)~\cite{ganin2015unsupervised} based adversarial classifier to eliminate speaker information in latent space for voice conversion tasks, extracting features related to speaker identity using a common classifier for timbre. In our work, we adopt the adversarial paradigm to decouple video features into actional and spatial components, inspired by these prior works.
 
\section{Methodology}
In this section, we formally introduce \textbf{DLL}, the method that decouples labels in both pattern level and knowledge level with the goal of debiasing the scene graph generation. Firstly, we present the preliminaries in Sec.\ref{sec:pf}. Secondly, we detail Pattern Decoupling Learning (PDL) in Sec.\ref{sec:fd}, which decouples predicate labels into patterns, transforming the original predicate prediction into a less biased pattern-level classification problem. Thirdly, we present Knowledge Decoupling Learning (KDL) in Sec.\ref{sec:ld}, which further boosts the tail performance by decoupling predicate knowledge into target and non-target knowledge and then calibrating the non-target knowledge of tail predicates using head classes within the same pattern. Finally, we describe the overall training objective of DLL in Sec.~\ref{sec:lo}.

\begin{figure}
\begin{center}
\includegraphics[width=\linewidth]{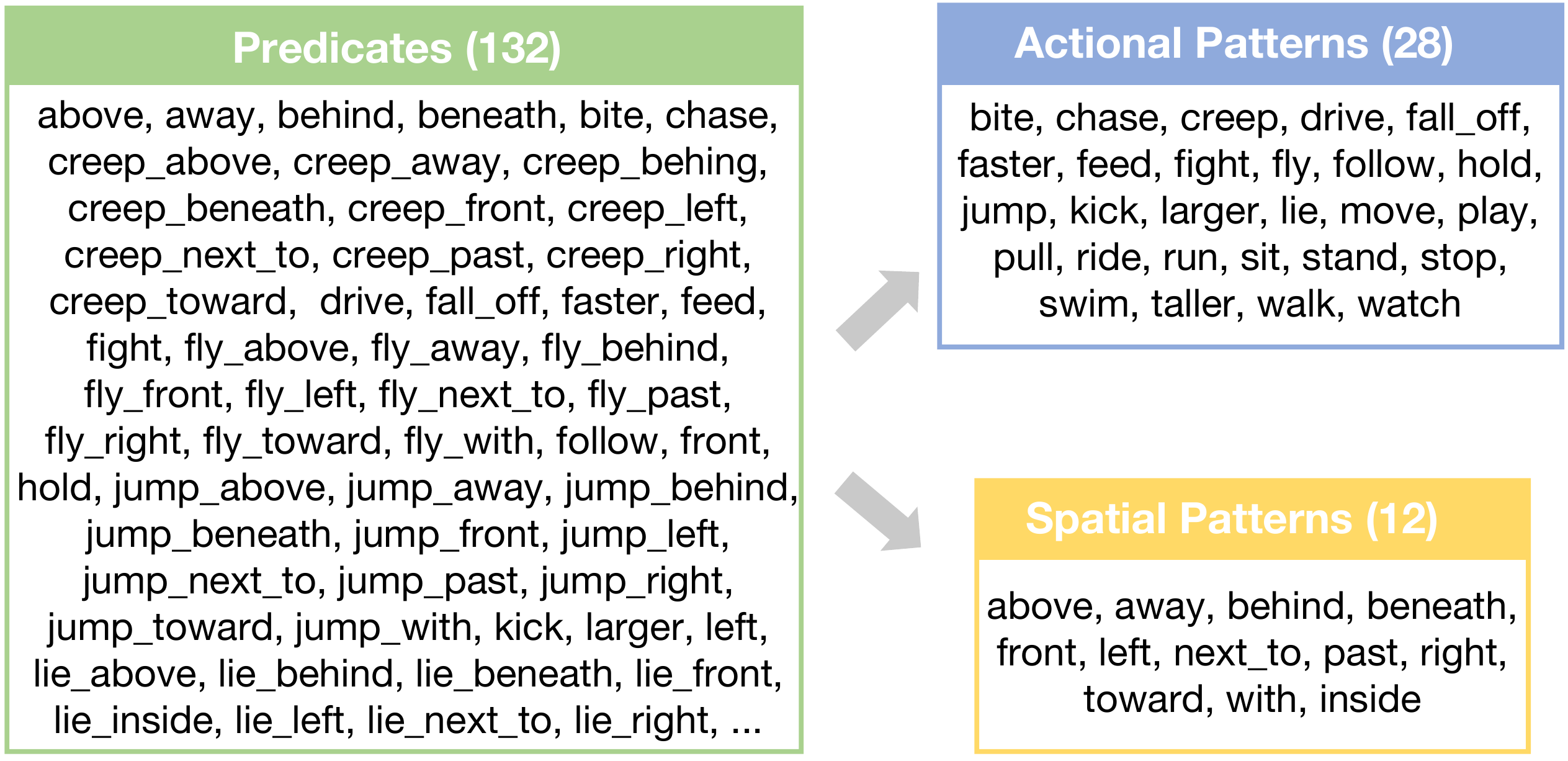} 
\end{center}
    \caption{Illustration of predicate labels and actional \& spatial patterns after label decoupling.}
    \label{fig:Pre2Pat}
\vspace{-0.2cm}
 \end{figure}

\begin{figure*}
\begin{center}
\includegraphics[width=\linewidth]{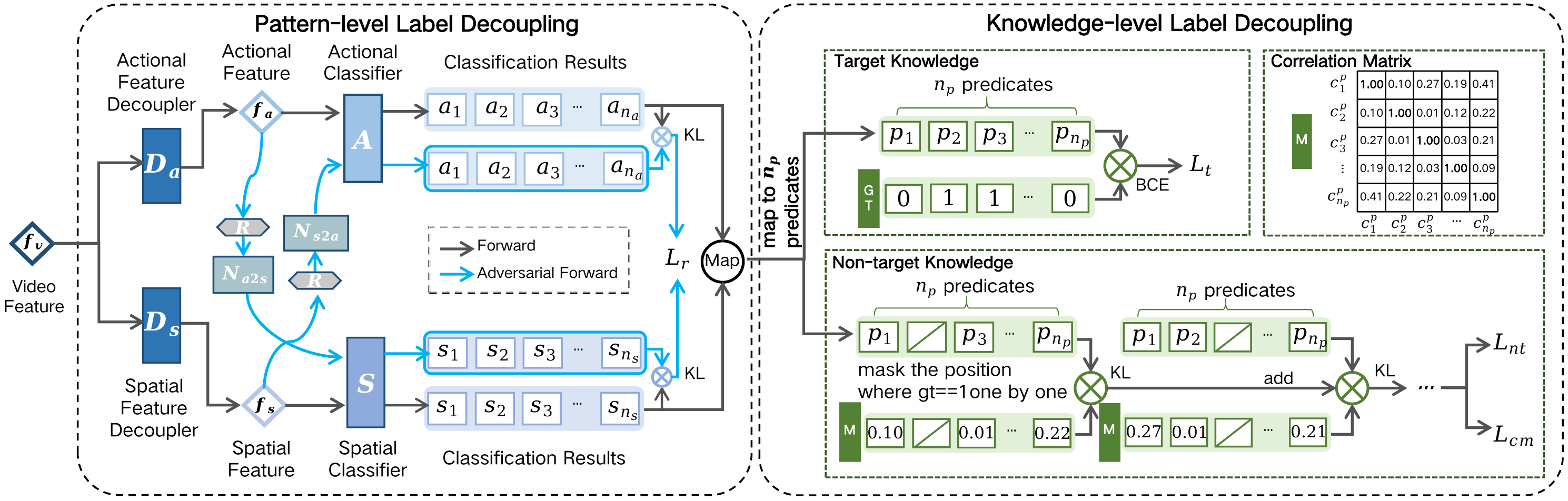} 
\end{center}
    \caption{\textbf{The overall framework of DLL.} DLL is fed with the video feature and conducts Pattern Decoupling Learning and Knowledge Decoupling Learning on the feature in a sequential manner.}
    \label{fig:main}
\vspace{-0.3cm}
 \end{figure*}

\subsection{Preliminary}\label{sec:pf}

\noindent\textbf{Problem Formulation.} Given an entity category set $\mathcal{C}_e$ and predicate category set $\mathcal{C}_p$, a video scene graph can be represented as $\mathcal{G} = (\mathcal{N},\mathcal{E})$, where $\mathcal{N}$ and $\mathcal{E}$ are the sets of nodes and edges, respectively. Each node in $\mathcal{N}$ is characterized by an entity category $c^e_i \in \mathcal{C}_e$ and a bounding box (bbox) sequence (tracklet). Each edge in $\mathcal{E}$ is characterized by the linkage from the $i$-th node (subject) to the $j$-th node (object), and a collection of multiple predicate categories $\mathcal{P}_{ij}=\{c_k^p \in \mathcal{C}_p | k=1,\ldots,n_{ij}\}$ where $n_{ij}$ is the total number of predicates between $i$-th and $j$-th node. 

\noindent\textbf{Actional and Spatial Pattern Sets.} 
We depict the actional and spatial pattern sets decoupled from predicate labels in Fig.~\ref{fig:Pre2Pat}. These exemplar predicate labels are from VidVRD~\cite{shang2017video} dataset.

\noindent\textbf{Pipeline Overview.} 
Given a video segment, we first detect bbox in each frame using FasterRCNN~\cite{ren2015faster} and apply Seq-NMS~\cite{han2016seq} to generate tracklet (\ie, bbox sequence). For relation classification, we combine the RoI Aligned visual feature of tracklet regions and the relative position feature of subject-object pairs as the video feature $f_v$ of this segment. The video feature $f_v$ is then forwarded to the pattern decoupling learning (PDL) module to predict its actional and spatial patterns. After that, the actional and spatial patterns are integrated and mapped back to predicate space. Finally, the knowledge decoupling learning (KDL) module is employed to further calibrate the tail predicate distribution. The detailed architecture of our DLL (PDL + KDL) is depicted in Fig.~\ref{fig:main}.

\subsection{Pattern-level Label Decoupling}\label{sec:fd}
The proposed PDL module takes the feature vector $f_v$ as input. As a video feature, $f_v$ consists of intertwined actional and spatial information. To decouple the highly entangled $f_v$ to pattern space, we introduce two decouplers $D_a, D_s$ to separate $f_v$ into actional feature $f_a$ and spatial feature $f_s$ using adversarial disentangle training. 

\noindent\textbf{Adversarial Feature Disentanglement.}
As analyzed above, the aim of decouplers $D_a, D_s$ is to decouple feature $f_v$ into actional feature $f_a$ and spatial feature $f_s$, which are then utilized to train two separate pattern classifier $A$ and $S$. The specific output is as follows:

\begin{equation}\label{eqn:bce2}
\begin{aligned}
 f_a &= D_a(f_v) &\quad f_s &= D_s(f_v)\\
 \bm{p_a} &= A(f_a) & \bm{p_s} &= S(f_s),
\end{aligned}
\end{equation}
where $\bm{p_a}$ and $\bm{p_s}$ is the output of $A$ and $S$, respectively.

However, to decouple the highly entangled feature $f_v$ to patterns is non-trivial, especially for those less supervised predicates. To tackle this challenge, we propose an adversarial learning method to achieve better disentanglement in a self-supervised manner: we design two extra actional-spatial and spatial-actional feature extraction network $N_{a2s}, N_{s2a}$ to learn opposite pattern features $f_{a2s}, f_{s2a}$ from $f_{s}, f_{a}$. Generally, if $f_a$ and $f_s$ are well disentangled, $f_{a2s}$ and $f_{s2a}$ should not contain any discriminative information about the opposite patterns. In this way, the specific output is as follows:

\begin{equation}\label{eqn:bce2}
\begin{aligned}
 f_{a2s} &= N_{a2s}(f_{a})  &\quad f_{s2a} &= N_{s2a}(f_{s}) \\  
 \bm{p_{s2a}} &= A(f_{s2a}) & \bm{p_{a2s}} &= S(f_{a2s}),
\end{aligned}
\end{equation}
where $\bm{p_{s2a}}$ and $\bm{p_{a2s}}$ is the output of $A$ and $S$, respectively.

We train the disentanglers $D_a, D_s$ by minimizing the opposite pattern information in $f_{a2s}, f_{s2a}$. Accordingly, the network parameter $\theta_{N_{*}},\theta_{D_{*}}$ will play a two-player mini-max game with the following objective function $\mathcal{L}_r(\theta_{N_*}, \theta_{D_*})$:
\begin{equation}
\begin{split}
\min\limits_{\theta_{N_{a2s}}}\max\limits_{\theta_{D_a}}\mathcal{L}_r(\theta_{N_{a2s}}, \theta_{D_a}),\\
\min\limits_{\theta_{N_{s2a}}}\max\limits_{\theta_{D_s}}\mathcal{L}_r(\theta_{N_{s2a}}, \theta_{D_s}).
\end{split}\label{eq:minmax}
\end{equation}

We use KL loss with a non-parametric Gradient Reversal Layer (GRL)~\cite{ganin2015unsupervised} as $R$ to achieve Eq.~\ref{eq:minmax}:
\begin{equation}
\begin{split}
    \mathcal{L}_{PLD} &= \mathcal{L}_{r}\\
     &= -\lambda (D_{KL}(\bm{p_{a2s}}||\bm{p_s})+D_{KL}(\bm{p_{s2a}}||\bm{p_a})),
\end{split}\label{eq:grl}
\end{equation}
where $\lambda$ is a hyper-parameter, $\bm{p_{a2s}}$ \& $\bm{p_{s2a}}$ are the output of $S$ \& $A$ using feature $f_{a2s}$ \& $f_{s2a}$. $\bm{p_a}$ \& $\bm{p_s}$ are the gradient-blocked outputs of $A$ \& $S$ using feature $f_{a}$ and $f_{s}$, respectively.
Following the training rule of GRL, $R$ will multiply the incoming gradient by a negative value, so that the training objectives of the network before and after $R$ are opposite. $R$ is allocated between the feature extractor $f_*$ and the opposite pattern learning network module $N_*$, as shown in Fig.~\ref{fig:main}. The actional-spatial and spatial-actional learning network parameters $\theta_{N_*}$ and disentangler parameter $\theta_{D_*}$ are updated as follows:
\begin{equation}
    \begin{split}
    \theta_{N_*} &\leftarrow \theta_{N_*} - \mu \frac{\partial\mathcal{L}_{r}}{\partial\theta_{N_*}}, \\
    \theta_{D_*} &\leftarrow \theta_{D_*} + \mu \frac{\partial\mathcal{L}_{r}}{\partial\theta_{D_*}},\label{eq:theta_update}
    \end{split}
\end{equation}
where $\mu$ is the learning rate, $\theta_{N_{*}}\in\{\theta_{N_{a2s}},\theta_{N_{s2a}}\}$ and  $\theta_{D_{*}}\in\{\theta_{D_{a}},\theta_{D_{s}}\}$.

\noindent\textbf{Mapping Function.}  In this part, we detail the mechanism to map actional pattern and spatial pattern back to the predicate prediction. Specifically, we couple the actional logits and spatial logits by $Map(.,.)$:
\begin{equation}
\begin{split}
    \bm{p} = Map(\bm{p_a},\bm{p_s}),
\end{split}\label{eq:pas}
\end{equation}
where $\bm{p}$ denotes the probability of final coupling labels and $Map(.,.)$ denotes the mapping function from actional \& spatial space to predicate space. Since the predicate-pattern correspondences are fixed and known, $Map(.,.)$ can be regarded as an inverse process of Fig.~\ref{fig:Pre2Pat}. For dual-pattern predicates such as \texttt{fly\_away}, $Map(.,.)$ calculates its probability by average the corresponding actional and spatial pattern logits. While for those single-pattern predicates such as \texttt{bite} and \texttt{inside}, which only belong to either actional pattern or spatial pattern, $Map(.,.)$ would directly assign the output logits of classifiers $A$ or $S$ to it. 

\noindent\textbf{Mutual Calibration.} In certain scenarios, classifiers $A$ and $S$ may generate high-probability pattern combinations that do not correspond to any predicate labels in the original label space, \eg, \texttt{bite} + \texttt{away}. To enable these classifiers to better align with the data distribution of the final predicate labels, we propose an iterative mechanism for mutual calibration (MC) of pattern-level and predicate-level distributions as follows:
\begin{equation}
\begin{split}
    (\bm{p_a'},\bm{p_s'}) = Map^{-1}(Map(\bm{p_a},\bm{p_s})),\label{eq:mm}
\end{split}
\end{equation}
\begin{equation}
\begin{split}
    \bm{p_a}&\leftarrow \eta\bm{p_a} + (1-\eta)\bm{p_a'}\\
    \bm{p_s}&\leftarrow \eta\bm{p_s} + (1-\eta)\bm{p_s'}.\label{eq:as}
\end{split}
\end{equation}

MC enhances the adaptability of pattern classifiers by iteratively updating the distribution of patterns and predicates. By refining the classification outputs at both levels, our proposed mechanism helps the patterns generated by the classifiers to align with the desired predicate labels. 

\begin{figure}[t]
\begin{center}
\includegraphics[width=0.8\linewidth]{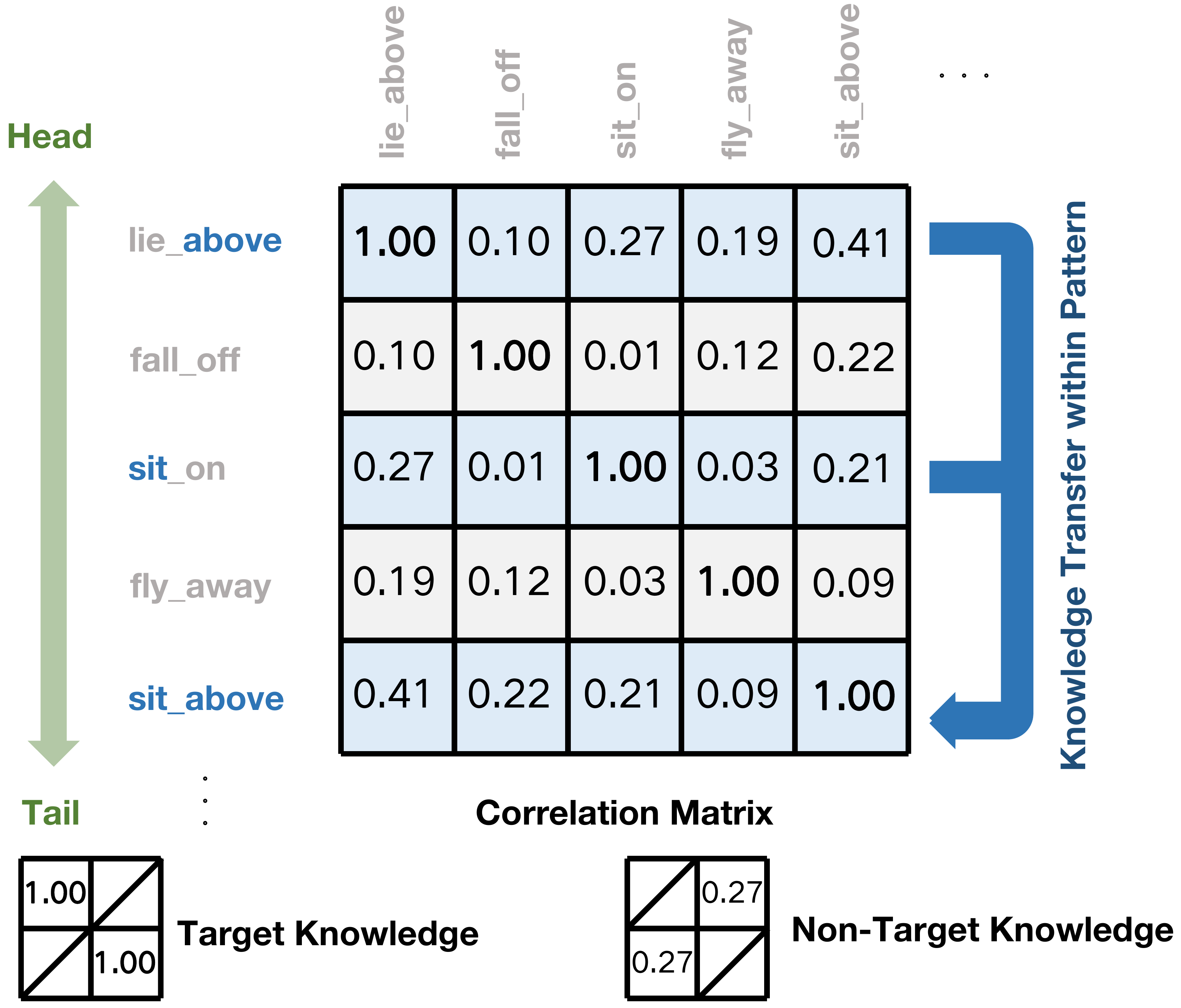} 
\end{center}
    \caption{Illustration of \textbf{KDL}, which transfers the non-target knowledge of head classes to the tail ones that are within the same actional or spatial pattern, in order to calibrate the biased distribution of tail predicates.}
    \label{fig:KDL}
\vspace{-0.2cm}
 \end{figure}
 
\begin{algorithm}[t]
\SetAlgoLined
\KwIn{disentangler $D_a, D_s$, actional-spatial and spatial-actional feature extraction network $N_{a2s}, N_{s2a}$, actional classifier $A$, spatial classifier $S$, correlation matrix $\bm{M}$.}
initialization: $\bm{M} \leftarrow \bm{I}, \alpha = 0.1, \gamma = 1.0$
\\
\For{1:epoch}{
   \For{1:iteration}{
       \textbf{Pattern-level Label Decoupling:}
        \\
        \For{1:step}{
            Update $\bm{p_a},\bm{p_s}$ using Eq.~\ref{eq:mm}, ~\ref{eq:as}\;
        }
        Update $D_a, D_s, A, S$ using Eq.~\ref{eq:bce}\;
        Update $D_a, D_s, N_{a2s}, N_{s2a}$ using Eq.~\ref{eq:grl} with $R$\;
        \textbf{Knowledge-level Label Decoupling:}
        \\
        Update $\bm{M}$ using Eq.~\ref{eq:matrix}\;
        Update $D_a, D_s, A, S$ using Eq.~\ref{eq:correlation}\;
        Update $\alpha$\;
        }
        Update $\gamma$\;
    }
 \caption{Decoupled Label Learning.}
\end{algorithm}

\subsection{Knowledge-level Label Decoupling}\label{sec:ld}
After acquiring the preliminary predicates, our next objective is to refine the distribution of the tail predicates with the assistance of the well-represented head ones. To achieve this, we divide the predicate label knowledge into two parts, namely target and non-target, and transfer the non-target knowledge from the head predicates to the tail ones that are within the same pattern, as shown in Fig.~\ref{fig:KDL}. 

\noindent\textbf{Target knowledge.} Target knowledge refers to the categorical information of a sample that represents the ground truth and is specific to each class. In the context of multi-label learning task such as VidSGG, we employ the binary cross-entropy loss function to supervise the target knowledge.
\begin{equation}
\begin{split}
    \mathcal{L}_{t} = BCE(\bm{p},\bm{q}),
\end{split}\label{eq:bce}
\end{equation}
where $\bm{p}$ denotes the prediction probabilities and $\bm{q}$ denotes the multi-hot encoded target labels.

\noindent\textbf{Non-target knowledge.} Non-target knowledge, also known as ``dark knowledge''~\cite{hinton2015distilling}, contains valuable information that represents the intrinsic correlation between classes. To capture this correlation, we utilize a correlation matrix denoted as $\bm{M} \in \mathbb{R}^{n_{p}\times{n_{p}}}$, where $n_p$ is the number of predicate labels. In our approach, we initialize the correlation matrix $\bm{M}$ as an identity matrix $\bm{I}$ and update it through mutual learning between the backbone model and $\bm{M}$.

\noindent\textbf{Correlation Matrix Updating.} We learn non-target knowledge from the model prediction probabilities to update the correlation matrix. For a sample of $k$-th predicate class with the probability vector $\bm{p}$, we employ KL loss between $k$-th position masked correlation vector $\bm{m}^{k}_{non}$ and gradient-blocked prediction probabilities $\bm{p}_{non}$ to update $\bm{M}$:
\begin{equation}
    \mathcal{L}_{cm} = D_{KL}(\bm{m}^{k}_{non}||\bm{p}_{non}) .
    \label{eq:matrix}
\end{equation}

\noindent\textbf{Head-to-Tail Knowledge Transfer.} We assume that the non-target knowledge of head predicates in $\bm{M}$ is well represented. To leverage the head knowledge, for a preliminary predicate, we search for the non-target knowledge of \emph{header} predicates in $\bm{M}$ that are within the same pattern, as presented in Fig.~\ref{fig:KDL}. The pinpointed knowledge is denoted as $\bm{m_h}$, where $h$ denotes the index of header predicate. Then we mask the $k$-th ground-truth position of $\bm{p}$ and $\bm{m_h}$ to learn the non-target knowledge with KL loss as follows:
\begin{equation}
    \mathcal{L}_{nt} = D_{KL}(\bm{p}_{non}||\bm{m}^{h}_{non}),
    \label{eq:correlation}
\end{equation}
where $k$ represents the existing target class index, $\bm{p}_{non}$ is prediction probabilities with $k$-th position masked, $\bm{m}^{h}_{non}$ is the masked gradient-blocked correlation vector $\bm{m_h}$, which denotes the $h$-th row in $\bm{M}$.
Since VidSGG is a multi-label task, such knowledge transfer would be continually proceeded until all the ground-truth positions are traversed.

The ``Correlation Updating'' and ``Knowledge Transfer'' form a mutual learning mechanism, enabling the VidSGG model and the correlation matrix $\bm{M}$ to learn from each other. Considering that the model learning is more reliable under the constraints of BCE loss in the early stage, we assign loss $\mathcal{L}_{nt}$ with a growing learning rate $\alpha$, which is initialized to a small value. The growth of $\alpha$ depends on the training iterations $i$ with hyper-parameter $\beta$. Accordingly, the overall loss of KDL is
\begin{equation}
    \begin{split}  
    \alpha &\leftarrow \alpha + \beta i, \\
    \mathcal{L}_{KDL} &= \mathcal{L}_{t} + \mathcal{L}_{cm} + \alpha\mathcal{L}_{nt} .\label{eq:ld_loss}
    \end{split}
\end{equation}

\subsection{Training Objective}\label{sec:lo}
With the above loss terms, the overall loss function of DLL can be written as:
\begin{equation}
    \mathcal{L} = \mathcal{L}_{PDL} + \gamma\mathcal{L}_{KDL},
\end{equation}
where $\gamma$ is employed to control the relative scale of the two loss terms.

\begin{table*}[t]
\centering
\caption{Performance (\%) on VidVRD~\cite{shang2017video} dataset. \textbf{Mean}: The average of mR@50/100 and R@50/100.} 
\label{tabel:vidvrd-sgdet}
\vspace{0.1cm}
\resizebox{0.9\linewidth}{!}{
    \begin{tabu}{l|cccccc|cc}
    \tabucline[2pt]{-}
     & \multicolumn{6}{c|}{Relation Detection}            & \multicolumn{2}{c}{Relation Tagging} \\ \multicolumn{1}{c|}{Method} & \multicolumn{1}{c}{mR@50} & \multicolumn{1}{c|}{mR@100} & R@50  & R@100 & \multicolumn{1}{|c|}{Mean} & \multicolumn{1}{c|}{mAP}      & P@5        & P@10       \\
    \tabucline[1.5pt]{-}
    VidVRD~\cite{shang2017video} $_{ACM-MM 2017}$                 & -     & \multicolumn{1}{c|}{-}      & 5.54  & 6.37  & \multicolumn{1}{|c|}{-}    & 8.58  & 28.90      & 20.80      \\
    GSTEG~\cite{tsai2019video}  $_{CVPR2019}$                & -     & \multicolumn{1}{c|}{-}      & 7.05  & 8.67  & \multicolumn{1}{|c|}{-}    & 9.52  & 39.50      & 28.23      \\
    3DRN~\cite{20213}  $_{Neurocomputing2021}$                  & -     & \multicolumn{1}{c|}{-}      & 5.53  & 6.39  & \multicolumn{1}{|c|}{-}    & 14.68  & 41.80      & 29.15      \\
    VRD-GCN~\cite{qian2019video}  $_{ACM-MM2019}$      & -     & \multicolumn{1}{c|}{-}      & 8.07  & 9.33  & \multicolumn{1}{|c|}{-}    & 16.26  & 41.00      & 28.50      \\
    MHA~\cite{su2020video}  $_{ACM-MM2020}$           & -     & \multicolumn{1}{c|}{-}      & 9.53  & 10.38 & \multicolumn{1}{|c|}{-}    & 19.03  & 41.40      & 29.45      \\
    TRACE~\cite{teng2021target}   $_{ICCV2021}$                & 7.55  & \multicolumn{1}{c|}{9.37}   & 9.08  & 11.15 & \multicolumn{1}{|c|}{9.29}    & 17.57  & 45.30      & 33.50      \\
    TSPN~\cite{woo2021and}   $_{Arxiv 2022}$                 & -     & \multicolumn{1}{c|}{-}      & 11.56 & 14.13 & \multicolumn{1}{|c|}{-}    & 18.90 & 43.80      & 33.73      \\
    Social Fabric~\cite{Chen2021Social} $_{ICCV2021}$         & -     & \multicolumn{1}{c|}{-}      & 13.73 & 16.88 & \multicolumn{1}{|c|}{-}    & 20.08  & 49.20      & 38.45      \\
    IVRD~\cite{li2021interventional}  $_{ACM-MM2021}$                  & -     & \multicolumn{1}{c|}{-}      & 12.40 & 14.46 & \multicolumn{1}{|c|}{-}    & 22.97  & 49.87      & 35.75      \\
    VRD-STGC+MSVGG~\cite{xu2022meta}  $_{ECCV2022}$              & -  & \multicolumn{1}{c|}{-}  & 12.62 & 15.78 & \multicolumn{1}{|c|}{-}    & 20.76   & 44.90      & 33.15  \\
    \tabucline[1.5pt]{-}
    VRD-STGC~\cite{liu2020beyond}  $_{CVPR2020}$              & \textbf{8.73}  & \multicolumn{1}{c|}{10.21}  & 11.21 & 13.69 & \multicolumn{1}{|c|}{10.96}    & \underline{18.38}   & \underline{43.10}      & 32.24  \\
    \textbf{+PDL (ours) }               & 7.75 & \multicolumn{1}{c|}{\textbf{10.76}}  & \textbf{12.43} & \underline{16.09} & \multicolumn{1}{|c|}{\textbf{11.76}}    & \textbf{18.40} & 41.40     & \textbf{32.31}  \\ 
    \textbf{+KDL (ours) }              & \underline{8.59} & \multicolumn{1}{c|}{10.52}  & \underline{12.41} & 15.47 & \multicolumn{1}{|c|}{\underline{11.75}}    & 18.30   & \textbf{43.50}      & 31.80      \\
    \textbf{+DLL (ours)}               & 7.85 & \multicolumn{1}{c|}{\underline{10.54}}  & \underline{12.41} & \textbf{16.15} & \multicolumn{1}{|c|}{11.74}    & 18.33  & 41.60      & \underline{32.25}     \\ 
    \tabucline[1.5pt]{-}
    VidVRD-II~\cite{shang2021video}   $_{ACM-MM2021}$           & 12.41 & \multicolumn{1}{c|}{12.97}  & 13.63 & 14.85 & \multicolumn{1}{|c|}{13.47}    & 25.93   & \underline{55.60}      & \underline{41.70}      \\ 
    \textbf{+PDL (ours) }               & \underline{13.09} & \multicolumn{1}{c|}{\underline{14.12}}  & 13.80 & \underline{15.39} &  \multicolumn{1}{|c|}{\underline{14.10}}    & 25.93 & \textbf{55.70}     & {41.30}     \\
    \textbf{+KDL (ours) }              & 12.23 & \multicolumn{1}{c|}{13.17}  & \underline{13.82} & {15.26} & \multicolumn{1}{|c|}{13.62}    & \underline{26.02} & 54.90      & \textbf{41.80}      \\
    \textbf{+DLL (ours)}               & \textbf{13.28} & \multicolumn{1}{c|}{\textbf{14.33}}  & \textbf{14.13} & \textbf{15.62} & \multicolumn{1}{|c|}{\textbf{14.34}}    & {\textbf{26.65}}  & {53.70}     & {41.20}     \\ 
    \tabucline[2pt]{-}
\end{tabu}
}
\vspace{-0.2cm}
\end{table*}

\begin{figure*}[t]
\centering
\subfigure[  \texttt{above}-related in VidVRD-II]{
\label{matrix.a}
\includegraphics[width=0.23\linewidth]{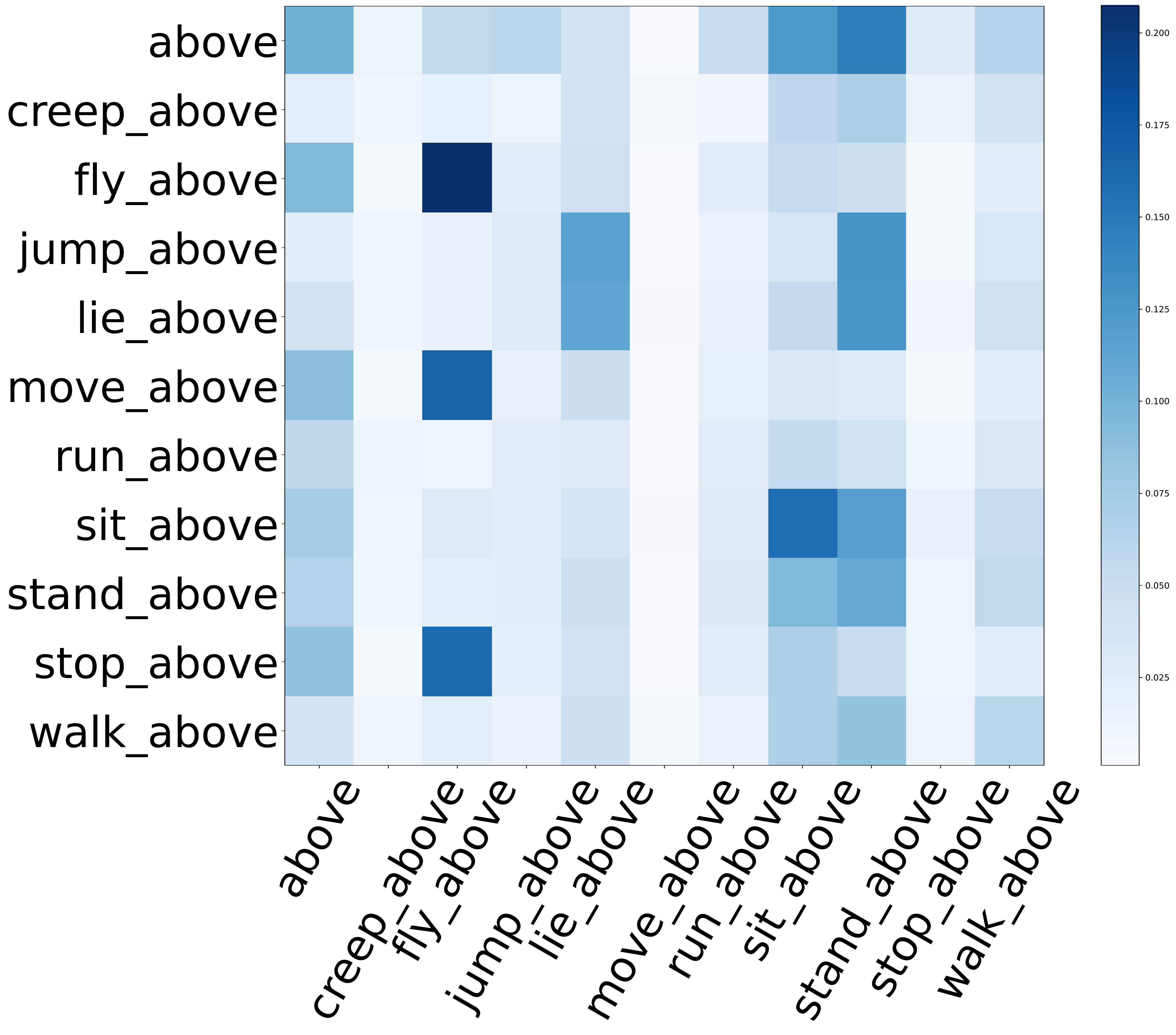}}
\subfigure[  \texttt{above}-related in \textbf{DLL (ours)}]{
\label{matrix.b}
\includegraphics[width=0.23\linewidth]{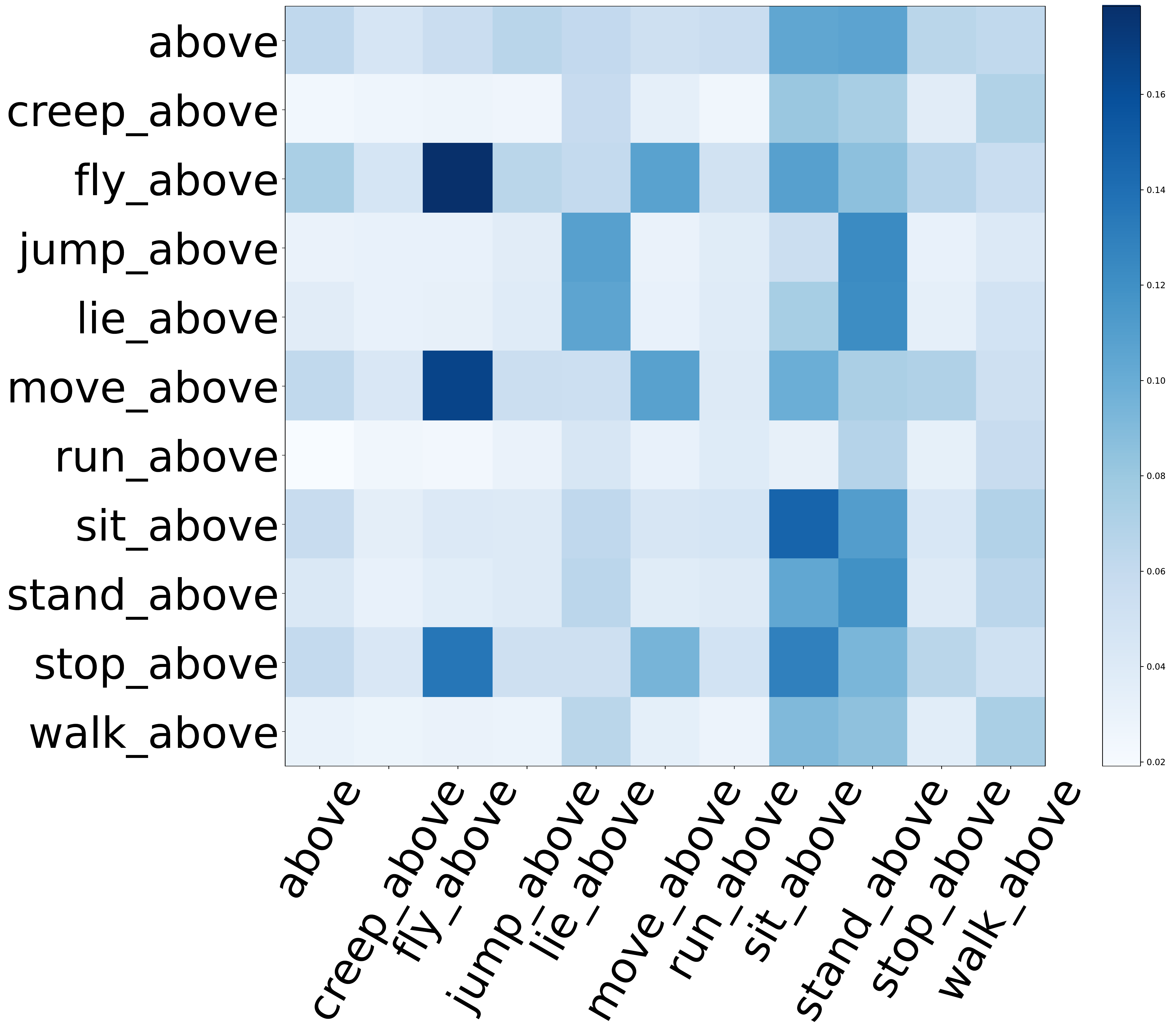}}
\subfigure[  \texttt{jump}-related in VidVRD-II]{
\label{matrix.c}
\includegraphics[width=0.23\linewidth]{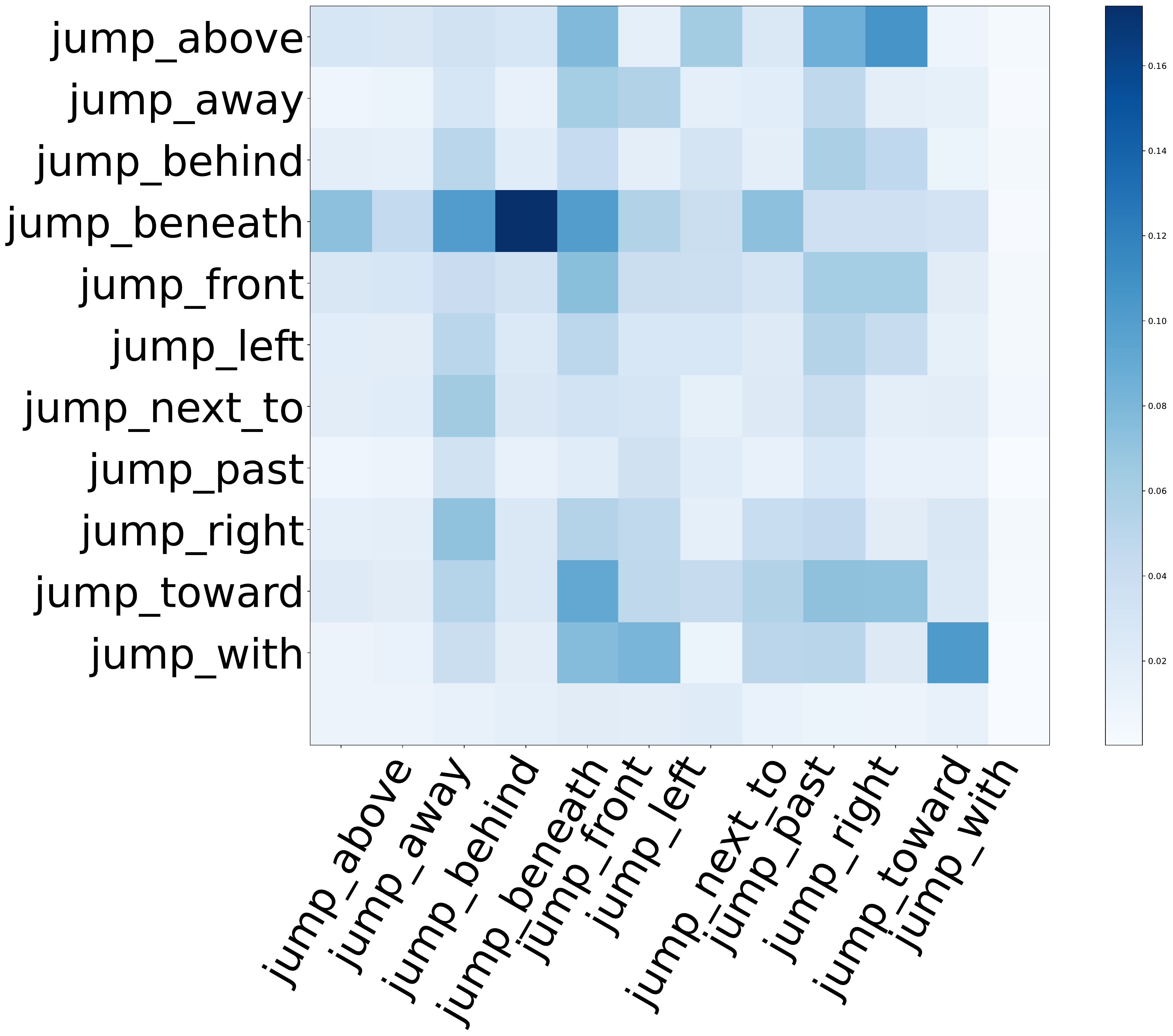}}
\subfigure[  \texttt{jump}-related in \textbf{DLL (ours)}]{
\label{matrix.d}
\includegraphics[width=0.23\linewidth]{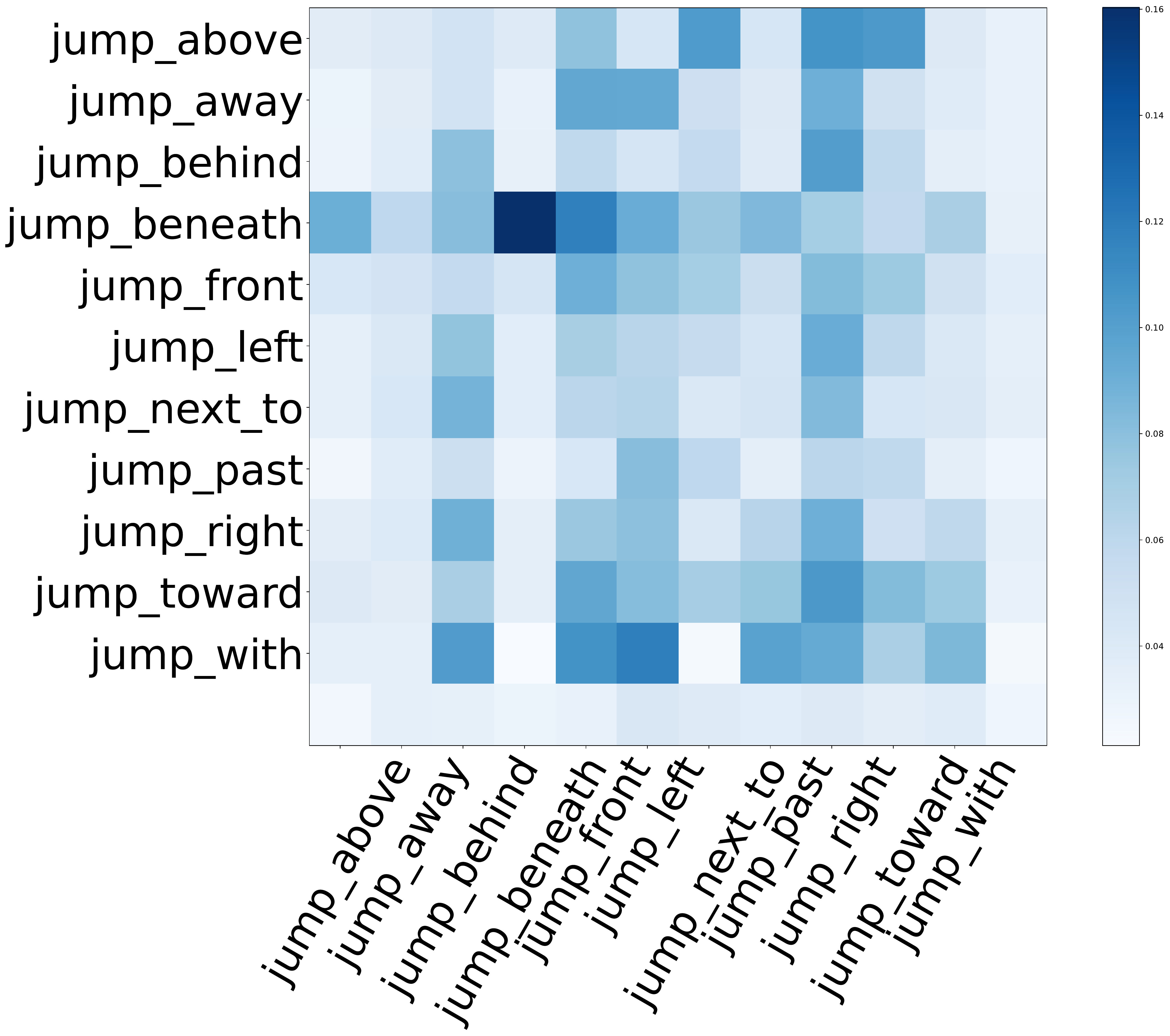}}
\caption{\textbf{Visualization of the prediction logits of the visual relations within same actional or spatial pattern.} We normalize the scores over the whole map and only display a part of the predicates. As can be seen, DLL can promote the correlations between same-pattern predicates thus transfers the well represented head knowledge to tail ones that are within the same pattern.}
\label{fig:matrix}
\vspace{-0.3cm}
\end{figure*}

\section{Experiments}

\subsection{Datasets}
We adopt the video relation benchmark ImageNet-VidVRD~\cite{shang2017video} in our experiments: 
\noindent\textbf{VidVRD~\cite{shang2017video}} is the first dataset for benchmarking VidSGG and has been widely used in previous work~\cite{liu2020beyond, teng2021target, Chen2021Social, woo2021and}. It contains $1,000$ videos ($800$ for training and $200$ for evaluation). 

\subsection{Evaluation Metrics}
\noindent\textbf{Metrics.} We use the official evaluation metrics~\cite{shang2017video,shang2021video} of the VRU Challenge, including Relation Detection (RelDet) and Relation Tagging (RelTag). Besides the average precision \textbf{(mAP)}, Recall@K (\textbf{R@K},K=$50,100$) for RelDet and Precision@K (\textbf{P@K},K=$5,10$) for RelTag. Following~\cite{chen2019knowledge,tang2019learning,tang2020unbiased} and~\cite{li2022label}, we first introduce the mean Recall@K \textbf{(mR@K)} and a comprehensive metric \textbf{Mean} as key metrics to VidSGG. 

\noindent\textbf{mR@K.} mR@K is calculated by obtaining recall scores from the top-K triplet predictions in every video segment first and then averaging them \emph{w.r.t} each predicate category. As discussed in ~\cite{chen2019knowledge,tang2019learning,tang2020unbiased}, mR@K serves as a more canonical metric in the long-tailed learning scenario. 

\noindent\textbf{Mean.} As discussed in ~\cite{li2022label}, Mean is the average of mR@K and R@K. Since R@K favors head predicates and mR@K favors tail predicates, the Mean metric is better for evaluating the performance across all predicates. 

\subsection{Implementation Details}

\noindent\textbf{Relation Detection Details.} \textbf{VRD-STGC~\cite{liu2020beyond}:} We employ a Faster-RCNN~\cite{ren2015faster} model in video frames, then track frame-level detection results across the whole video using a Multiple Object Tracking (MOT) algorithm to obtain tracklets. We adopt RoI Aligned detection and I3D feature with relative motion features for each pair. Only pairs overlapping with the ground truth by more than $0.5$ in vIoU(volume IoU) are selected. Then we construct a spatial graph and a temporal graph to filter incompatible proposals. \textbf{VidVRD-II~\cite{shang2021video}:} Given a video, we first split it into shot segments of 30 frames with 15 frames overlapped, then detect the bboxes in each frame using Faster-RCNN~\cite{ren2015faster} and apply Seq-NMS~\cite{han2016seq} to generate tracklets (i.e., bounding box sequence). We adopt the RoI Aligned visual feature of tracklet regions and the relative position feature of subject-object pairs. Finally, we perform the simple greedy relation association as proposed in~\cite{shang2017video} to associate the detected relation instances across the segments.

\noindent\textbf{Hyper-paramaters}. \textbf{VRD-STGC~\cite{liu2020beyond}:} We set $\lambda$=1e-1, $\eta$=1e-1, $\beta$=1e-4 and $\gamma$ = $pow(0.99,e)$ where $e$ denotes the epoch. We set $\alpha$=1e-1 in the first $3$ epochs. In order to better couple labels, we fine-tune the ratio of actional labels and spatial labels. Other hyper-parameters are set consistently with VRD-STGC~\cite{shang2021video}. Our model is trained for total of $20$ epochs with the learning rate $\mu$=1e-1 by using SGD~\cite{1951A} optimizer. \textbf{VidVRD-II~\cite{shang2021video}:} We set $\lambda$=0.13, $\eta$=1e-3, $\beta$=1e-4 and $\gamma$ = $pow(0.99,e)$. We set $\alpha$=1e-1 in the first $10$ epochs. Other hyper-parameters are set consistently with VidVRD-II~\cite{shang2021video}. We train DLL for total of $100$ epochs with the learning rate $\mu$=1e-3 by using Adam~\cite{kingma2014adam} optimizer. 

\begin{figure*}[t]
\begin{center}
\includegraphics[width=\linewidth]{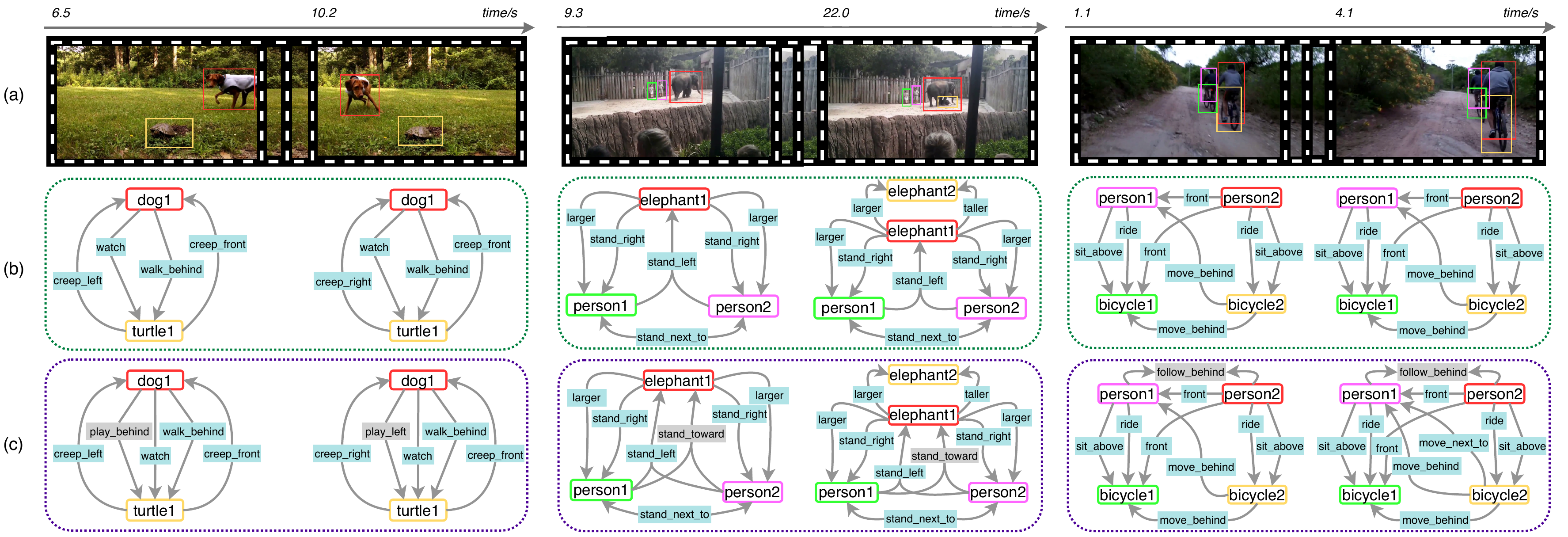} 
\end{center}
    \caption{\textbf{Visualization of scene graph generation results and open-vocabulary relation prediction.} (a) Results of object detection. (b) Results of predicted triplets by \textbf{DLL}. (c) Possible open-vocabulary predicates captured by DLL, which are marked with grey background.}
    \label{fig:vid}
\vspace{-0.3cm}
 \end{figure*}

\subsection{Comparison with State of the Arts}
\noindent\textbf{Performance Comparison on VidVRD.} In Table.~\ref{tabel:vidvrd-sgdet}, we evaluate our DLL by incorporating it into two typical baseline models in VidSGG: VRD-STGC~\cite{liu2020beyond} and VidVRD-II~\cite{shang2021video}. However, since other state-of-the-art methods' goals are not to solve the long-tail problem, they have not recorded their performance on mR metrics, so we do not have these relevant data. From Table.~\ref{tabel:vidvrd-sgdet}, we have the following observations: 1) Compared with the two baselines, our DLL consistently improves the model performance on both R@K and mR@K metrics, as well as the Mean and mAP. \emph{e.g.} DLL achieves a significant improvement (0.85\%/1.36\%) in mR@50/100 based on VidVRD-II and 0.33\% in mR@100 based on VRD-STGC. This demonstrates the effectiveness of DLL in dealing with the long-tail problem in predicate distribution.Unlike existing methods that have to trade-off between R@K and mR@K, DLL can improve both metrics simultaneously. This can be confirmed by the 1.21\%/2.46\% and 0.5\%/0.77\% improvement in R@50/100, 0.87\% and 0.78\% improvement in Mean compared to two baselines with a slight loss in P@5. 2) Our DLL method outperforms our single module methods PDL and KDL. According to our observation, we can see that KDL could enhance mR@K metric while PDL is more capable of solving the long-tail problem than KDL. Although both PDL and KDL may have some slight losses on mR@50, the combination of PDL and KDL can be complementary to each other and achieve synergistic improvement.

\subsection{Ablation Study}

\noindent\textbf{Effectiveness of the Adversarial Disentanglement and Mutual Calibration.} 
Table.~\ref{tabel:ablation_adversarial} shows the performance of PDL without the KDL module on VidVRD-II~\cite{shang2021video}, with or without adversarial disentanglement and mutual calibration. Compared with the baseline in row one, PDL with adversarial disentanglement can further improve the performance and enable the actional and spatial classifiers to learn the actional and spatial feature more effectively, \emph{e.g.} 0.69\%/0.73\% in mR@50/100 and 0.7\% in R@100. This indicates that adversarial disentanglement can make the classifier obtain pure and sufficient learning. Therefore, we adopt adversarial disentanglement in our later work. We also observed that MC is a good way to help the classifier align with the data distribution of the final predicate labels by row 4 of Table.~\ref{tabel:ablation_adversarial}, which is reflected in the overall improvement of performance.

\begin{table}[t]
\centering
\caption{Ablation (\%) of w/o Adversarial Disentanglement (AD) and Mutual Calibration (MC) in PDL without KDL based on VidVRD-II~\cite{shang2021video}.}
\label{tabel:ablation_adversarial}
\vspace{0.1cm}
\resizebox{\linewidth}{!}{
\begin{tabu}{cc|[1pt]c|c|c|c|c}
\tabucline[2pt]{-}
AD &  MC  & mR@50/100 & R@50/100 & Mean & mAP & P@5/10 \\
\tabucline[1.5pt]{-}
\multicolumn{2}{c|}{VidVRD-II}   & 12.41/12.97 & 13.63/14.85 & 13.47 & 25.93 & 55.60/\textbf{41.70} \\
\checkmark &  & 12.56/13.70 & 13.59/\textbf{15.55} &  13.88  & \textbf{26.10} & 54.50/41.25 \\
 & \checkmark & 12.26/13.09 & \textbf{13.80}/15.18 &  13.58  & 25.12 & 53.20/38.75 \\
\checkmark & \checkmark & \textbf{13.09/14.12} & \textbf{13.80}/15.39 &  \textbf{14.10}  & 25.93 & \textbf{55.70}/41.30\\
\tabucline[2pt]{-}
\end{tabu}
}
\vspace{-0.2cm}
\end{table}

\begin{table}[t]
\centering
\caption{Ablation (\%) of different steps of Mutual Calibration in PDL without KDL based on VidVRD-II~\cite{shang2021video}.}
\label{tabel:ablation_iterative}
\vspace{0.1cm}
\resizebox{\linewidth}{!}{
\begin{tabu}{c|[1pt]c|c|c|c|c}
\tabucline[2pt]{-}
step/$\eta$  & mR@50/100 & R@50/100 & Mean & mAP & P@5/10 \\
\tabucline[1.5pt]{-}
0/0  & 11.75/13.60 & 12.68/14.25 & 13.07 & 23.50 & 51.50/39.40 \\
\tabucline[1.0pt]{-}
1/1e-1  & 11.79/12.68 & 12.84/14.60 &  12.98  & 24.45 & 53.40/40.80 \\
\tabucline[1.0pt]{-}
\textbf{1/1e-2}  & \textbf{13.58/14.62} & 14.04/15.53 &  \textbf{14.44}  & \textbf{26.00} & \textbf{56.10}/40.80 \\
2/1e-2  & 12.23/13.13 & 13.22/14.62 &  13.30  & 24.04 & 54.30/39.95 \\
3/1e-2  & 12.95/14.23 & \textbf{14.21/15.82} &  14.30  & 25.59 & 54.30/\textbf{41.00} \\
\tabucline[1.0pt]{-}
1/1e-3  & 12.40/13.53 & 13.34/14.91 & 13.55 & 26.31 & 53.80/39.80 \\
\tabucline[2pt]{-}
\end{tabu}
}
\vspace{-0.3cm}
\end{table}

\noindent\textbf{Number of the Mutual Calibration Steps.} Table.~\ref{tabel:ablation_iterative} shows the performance of different steps of mutual calibration with different scales of hyper-parameter $\eta$. Compared with the baseline that has no mutual calibration (step=0), the model that iteratively calibrates once achieves better results. From the data in Table~\ref{tabel:ablation_iterative}, we can see that mutual calibration improves not only R@K, but also mR@K, mAP and Mean, which indicates that mutual calibration can facilitate label coupling. However, as the hyper-parameter $\eta$ and iteration steps increase, the performance decreases, which suggests that excessive calibration is detrimental.

\noindent\textbf{Growth Rate of Knowledge Transfer.} 
Table.~\ref{tabel:ablation_ld} shows the performance of using different $\beta$ in Knowledge Transfer without PDL module on VidVRD-II~\cite{shang2021video}. It can be seen that a certain growth rate can improve the performance and promote the mutual learning between the model and the correlation matrix $\bm{M}$, which also proves the effectiveness of the KDL module.

\subsection{Qualitative Results}
\noindent\textbf{Visualization of correlations of predicates within the same actional or spatial pattern.} Fig.~\ref{fig:matrix} compares the correlations of predicates generated by baseline and DLL. As can be seen, DLL can promote the correlations between the similar predicates via non-target knowledge calibration.

\noindent\textbf{Visualization of predictions in open-vocabulary scenario.} Fig.~\ref{fig:vid} shows some detected predicates of DLL. Comparing to
the vanilla scene graphs, our generated ones are more promising in predicting open-vocabulary predicates.

\begin{table}[t]
\centering
\caption{Ablation (\%) of different scales of $\beta$ in Knowledge Transfer without PDL based on VidVRD-II~\cite{shang2021video}.}
\vspace{0.1cm}
\label{tabel:ablation_ld}
\resizebox{\linewidth}{!}{
\begin{tabu}{c|[1pt]c|c|c|c|c}
\tabucline[2pt]{-}
$\beta$  & mR@50/100 & R@50/100 & Mean & mAP & P@5/10 \\
\tabucline[1.5pt]{-}
\textbf{1e-1}  & \textbf{12.23}/13.17 & \textbf{13.82/15.26 }&  \textbf{13.62}  & \textbf{26.02} & \textbf{54.90/41.80} \\
1e-2  & 12.10/12.99 & 13.69/14.97 &  13.44  & 25.68 & 54.30/40.85 \\
1e-3  & 11.99/\textbf{13.64} & 12.84/14.60 & 13.27  & 25.52 & 53.40/39.90 \\
1e-4  & 12.14/12.90 & 13.34/14.93 & 13.33  & 24.97 & 53.50/41.45 \\
\tabucline[2pt]{-}
\end{tabu}
}
\vspace{-0.3cm}
\end{table}

\section{Conclusion}
In this paper we target at VidSGG task, or more specifically, the long-tail problem inherently existed in the training data that hinders the VidSGG performance. To this end, we introduce \textbf{DLL}, a novel approach that decouples labels into actional and spatial patterns and learns them separately. DLL also transfers the non-target knowledge from head to tail predicates within the same pattern to further calibrate the tail predicate distribution. By combining these two decoupling method, we are able to create an ``unbiased'' scene graph, which accurately captures the visual relations in the video. Extensive results verify that DLL achieves state-of-the-art performance on various metrics across different scenarios, especially on tail predicates. Furthermore, the decoupled manner also improves the zero-shot learning ability of VidSGG model in open-vocabulary relation scenario.

{\small
\bibliographystyle{ieee_fullname}
\bibliography{egbib}
}

\end{document}